\documentclass[journal]{IEEEtran}

\usepackage{cite}
\usepackage{amsmath}
\usepackage{algorithmic}
\usepackage{url}
\usepackage{graphicx}
\usepackage{array}
\usepackage{amssymb}
\usepackage{multirow}
\usepackage[caption=false,font=footnotesize]{subfig}
\usepackage[pagebackref=true,breaklinks=true,colorlinks,bookmarks=false]{hyperref}
\usepackage{amsthm}
\usepackage{dsfont}
\usepackage{booktabs}
\usepackage{xcolor}
\usepackage[T1]{fontenc}
\usepackage{algorithmic}
\usepackage[ruled, vlined]{algorithm2e}
\usepackage{graphicx}
\usepackage{subfig}
\usepackage{booktabs}
\usepackage{threeparttable}
\usepackage{amssymb}
\usepackage{multicol}
\usepackage{multirow}
\usepackage{booktabs}
\usepackage{makecell}
\usepackage{hhline}
\usepackage{color}
\usepackage{float}
\usepackage[section]{placeins}

\SetKwInput{KwInput}{Input}             
\SetKwInput{KwOutput}{Output}
\graphicspath{{fig/}}

\begin{document}
\title{A Light-weight Transformer-based Self-supervised Matching Network for Heterogeneous Images}

\author{
	Wang Zhang,
	Tingting Li,
	Yuntian Zhang,
	Gensheng Pei,
	Xiruo Jiang,
    and~Yazhou Yao
	\thanks{W. Zhang, T. Li, Y. Zhang, G. Pei, X. Jiang, and~Y. Yao are with the School of Computer Science and Engineering, Nanjing University of Science and Technology, Nanjing, China.}
}

\markboth{}%
{Shell \MakeLowercase{\textit{et al.}}: I2CRC}

\maketitle

\begin{abstract}

Matching visible and near-infrared (NIR) images remains a significant challenge in remote sensing image fusion. The nonlinear radiometric differences between heterogeneous remote sensing images make the image matching task even more difficult. Deep learning has gained substantial attention in computer vision tasks in recent years. However, many methods rely on supervised learning and necessitate large amounts of annotated data. Nevertheless, annotated data is frequently limited in the field of remote sensing image matching. To address this challenge, this paper proposes a novel keypoint descriptor approach that obtains robust feature descriptors via a self-supervised matching network. A light-weight transformer network, termed as LTFormer, is designed to generate deep-level feature descriptors. Furthermore, we implement an innovative triplet loss function, LT Loss, to enhance the matching performance further. Our approach outperforms conventional hand-crafted local feature descriptors and proves equally competitive compared to state-of-the-art deep learning-based methods, even amidst the shortage of annotated data.
Code and pre-trained model are available at \textcolor{red}{https://github.com/NUST-Machine-Intelligence-Laboratory/LTFormer}.

\end{abstract}


\IEEEpeerreviewmaketitle

\section{Introduction}\label{sec:intro}

Heterogeneous remote sensing image matching pertains to the process of matching images from different sensors to achieve spatial congruence of homologous points under multi-modal imaging techniques. This task is crucial for various remote sensing applications such as land feature change detection \cite{Change_Detection,yao2023automated}, remote sensing image fusion \cite{imagefusion_2}, and target identification and tracking \cite{track}. With the continuous development of deep learning technologies \cite{sun2022pnp,NPN,mao2023attention,jiang2022deep}, multiple sensor technologies, remote sensing images can originate from different sensors, including visible light, infrared, and synthetic aperture radar. The images captured by these diverse sensors hold unique information and exhibit variations in imaging details for identical objects, providing complementarity. To garner richer image insights, matching the heterogeneous images is imperative before proceeding with fusion. However, this is challenged by significant radiometric and geometric disparities caused by variations in imaging mechanisms, band configurations, spatial resolution, spectral resolution, and temporal phase across different sensors. Consequently, tackling heterogeneous remote sensing image matching has become an intricate problem demanding further explorations and solutions in the domain of image matching. Some common examples of heterogeneous image pairs encompass visible-NIR (near-infrared), visible-infrared, optical-SAR, and electron computed tomography imaging-MRI, among others. In the present study, we narrow our focus specifically to the problem of matching visible-NIR heterogeneous remote sensing image pairs.

The current mainstream image matching methods can be roughly divided into two directions: area-based and feature-based matching methods. Area-based methods \cite{medical,imagefusion} strive to achieve high-precision image matching by leveraging pre-established similarity measures and transformation models. Such approaches primarily analyze superficial image information, such as grayscale or phase features, to facilitate direct pixel-level image matching. It establishes an intuitive and easily comprehendible direct mapping correlation between the superficial information and the transformation model to effectuate image matching. Nevertheless, area-based methods rely considerably on similarity measures, and factors like geometric variations, brightness fluctuations, and image distortions may lead to deceptive similarity measures. This challenges achieving high-precision matching for heterogeneous images with substantial differences. Furthermore, remote sensing images typically encompass larger spatial extents and are more susceptible to redundancy of shallow image information. Consequently, area-based remote sensing image matching methods frequently grapple with suboptimal real-time performance and elevated memory overhead.

On the other hand, feature-based image matching techniques \cite{Handcrafted,BFSIFT} accomplish the task by extracting geometric features like points, lines, and surfaces from the image, subsequently generating local feature information descriptors. Such methods capitalize on the image's sparse yet salient spatial geometric features, thereby achieving efficient image matching. They adeptly mitigate the issue of information redundancy by abstracting the original comprehensive image for representation. Consequently, feature-based image matching techniques are apt for tasks involving considerable image differences. The heterogeneous remote sensing image matching framework we have designed also employs this feature-based approach.

Hand-crafted feature descriptors, though extensively employed in numerous visual applications, frequently falter when confronting heterogeneous images due to the existence of nonlinear radiometric variations. In contrast, deep learning-based feature descriptors have shown superior capability in capturing deep image information and providing better feature descriptions. Our work concentrates on resolving the feature matching conundrum in visible and near-infrared remote sensing images. We introduce a self-supervised descriptor termed LTFormer, which is constructed upon the light-weight transformer architecture. Extensive experiments demonstrate the effectiveness of LTFormer in handling feature matching in visible and near-infrared remote sensing images. The main contributions of our work can be summarized as follows:

(1) We design a self-supervised training method that utilizes ternary data and propose a data construction strategy to facilitate the matching process.

(2) We develop a light-weight pyramid-based transformer network to generate deep feature descriptors, achieving comparable performance with fewer parameters.

(3) We instantiate the LT Loss function grounded on Triplet Loss, which decreases the distance between positive and anchor samples while increasing the distance from negative samples. This loss function amplifies the discriminative prowess of the feature descriptors.

The remaining sections of this paper are organized as follows: Section \ref{sec:rel_work} introduces the related work on image matching. Section \ref{sec:proposed} provides a detailed description of our method. Section \ref{sec:experiment} discusses the experiments and ablation results. The conclusion of this work is presented in Section \ref{sec:conclusions}.

\section{Related Work}\label{sec:rel_work}
\subsection{Area-based matching method}
Nowadays, the application of multi-modal images is increasingly widespread in various domains. For example, in the field of medical imaging, \cite{medical} propose a new unsupervised adaptive method for segmenting cardiac CT images and cardiac MRI images. In environmental monitoring, fusing visible images with infrared images \cite{imagefusion} can provide better object detection and tracking capabilities, improving security surveillance and resource management. Image matching plays a crucial role in enabling effective information fusion between multi-modal images. In the field of remote sensing image matching, numerous algorithms have emerged, which can be categorized into two main types: area-based matching methods and feature-based matching methods. Let's provide an overview of these two main categories: area-based and feature-based.

In traditional area-based matching methods, grayscale features of images have always been a focal point of attention. Commonly used similarity measurement methods based on grayscale features include Sum of Squared Differences (SSD), Normalized Cross-Correlation (NCC), Mutual Information (MI), \cite{SMI,MI,NCC}. These methods are relatively simple to implement but are sensitive to external factors such as noise and grayscale variations. Introducing deep learning for estimating image similarity and transformation models has become a popular direction in this category of methods. For remote sensing images that lack obvious features, \cite{GLI} propose a multi-source image matching algorithm based on geographic location information. This algorithm achieves image registration by computing the corresponding positions of different pixels at the same geographic location in two pairs of images, without being limited by image features and imaging platforms. To address the issue of poor localization accuracy of traditional feature extraction methods in remote sensing images, GMatchNet~\cite{GMatchNet} utilizes geometric CNN to estimate affine transformation parameters based on matching patches. It further refines the central coordinates to improve the localization accuracy of feature points.

Dense optical flow estimation is a special type of displacement field estimation method that can handle complex spatial mapping relationships. However, it is computationally challenging, inefficient, and has strict prerequisites. It has been commonly used in medical imaging and the registration of natural images with short time spans. Recently, deep learning-based dense optical flow estimation has also been applied to the matching problem of high spatial resolution remote sensing images. LiteFlowNet~\cite{LiteFlowNet} is designed as an alternative network to FlowNet2~\cite{FlowNet2}, achieving comparable performance while significantly reducing the number of parameters. It utilizes a pyramid feature extraction structure and proposes a more efficient flow inference method at each pyramid level using a light-weight cascade network, enabling dense optical flow estimation. LiteFlowNet2~\cite{LiteFlowNet2} and LiteFlowNet3~\cite{LiteFlowNet3} further improve the accuracy and efficiency of optical flow estimation. LiteFlowNet2 reduces the number of layers in the original image pyramid based on computational cost and accuracy improvement analysis. It also introduces a simple inference network called NetE, which effectively enhances computational efficiency and estimation accuracy. LiteFlowNet3 introduces a local flow consistency constraint to further improve the accuracy of optical flow estimation.

Overall, due to the larger spatial extent and complex geometric distortions present in remote sensing images, area-based matching methods are less commonly used in the field of remote sensing imagery.

\subsection{Feature-based matching methods}
\subsubsection{Handcrafted Local Feature Descriptors}
With the proposal and improvement of the floating descriptor~\cite{Handcrafted} SIFT~\cite{SIFT}, the feature-based image matching technique has gradually become a research hotspot. While SIFT ensures scale and rotation invariance through the construction of a Gaussian difference pyramid, it overlooks the rationality of feature point distribution. To filter out feature points with abnormal distributions, UR-SIFT~\cite{UR-SIFT} utilizes stability and uniqueness constraints to extract evenly distributed, reliable, and accurately aligned feature points. MUR-SIFT~\cite{MUR-SIFT} improves upon SIFT to obtain uniformly distributed matching features. PC-SIFT~\cite{PC-SIFT} combines the coarse matching results obtained by SIFT with the fine matching obtained through phase correlation. Considering the matching speed, \cite{SURF} propose SURF, which uses integral images and Haar wavelet transforms to generate descriptors, reducing computational costs and accelerating descriptor extraction. However, it exhibits inferior scale and rotation invariance compared to the SIFT algorithm.

While SIFT features based on the Gaussian scale space blur the natural boundaries of objects, KAZE~\cite{KAZE} constructs a nonlinear scale space and employs nonlinear diffusion filtering to describe features. However, it comes with a high computational cost. Similarly, BFSIFT~\cite{BFSIFT}, which utilizes bilateral filtering to construct an anisotropic scale space for feature detection, also suffers from expensive computational costs. Since the introduction of BRIEF~\cite{BRIEF}, the use of binary strings as efficient feature descriptors has become increasingly popular. In order to reduce the computational complexity while maintaining the performance of methods such as SIFT and SURF, binary descriptors such as BRISK~\cite{BRISK} and ORB~\cite{ORB} have been proposed, which are computationally two orders of magnitude faster than SIFT. AKAZE~\cite{AKAZE}, as an accelerated version of KAZE, embeds Fast Explicit Diffusion (FED) in the pyramid framework, which significantly improves the speed of feature detection in nonlinear scale spaces.

In the field of heterogeneous remote sensing image matching, \cite{EAT} propose an Enhanced Affine Transformation (EAT) method for non-rigid infrared and visible light image matching. The method transforms the image matching problem into a point set matching problem by extracting interest points from the image edge mapping. Then, they use the EAT model with optimal local feature estimation to construct the global deformation and simplify the Gaussian field-based objective function using potential correspondences between image pairs to guide the matching process.

\subsubsection{Deep Learning based Local Feature Descriptors}
In recent years, many studies have applied deep learning to feature-based image matching methods \cite{yao2017exploiting,tang2023holistic,yao2021non,pei2022eccv}. These methods generally embed deep learning models into the framework of traditional matching methods, replacing the feature extraction \cite{10298026,10128961}, description \cite{yao2020towards,10105896}, and matching stages \cite{yao2021jo,sun2021webly}. Compared to traditional methods, deep learning-based matching methods extract deep semantic information from images and learn complex spatial mapping relationships. This enables them to meet the requirements of efficient and accurate matching in remote sensing image matching.

The successful proposal of MatchNet~\cite{MatchNet}, a dual network architecture consisting of a feature extraction network and a learning metric network, opens a new chapter in the use of CNN for image matching. TFeat~\cite{TFeat} uses three sets of training samples to mine the positive and negative information to obtain better feature descriptors. L2-Net~\cite{L2Net} is a two-part network consisting of a feature detector and a feature descriptor. The detector network quickly extracts the information of the feature points in the image and inputs it into the descriptor network to obtain the feature descriptors. Based on the in-depth study of L2 regularisation, HyNet~\cite{HyNet} applies L2 regularisation to all feature mappings in the network, which further improves the performance of feature descriptors. In addition, the GeoDesc~\cite{GeoDesc} descriptor for multi-view reconstruction generated by integrating geometric constraints compensates for the lack of local descriptors in the image-based 3D reconstruction benchmarking and improves the matching accuracy.

SoSNet~\cite{SosNet} introduces Second-Order Similarity (SOS) to learn local descriptors and proposes Second-Order Similarity Regularisation (SOSR) method. The descriptors obtained by training perform well on some challenging tasks. DGD-net~\cite{DGDNet} uses VGG16 as a backbone network to extract dense feature descriptors. It constructs bootstrap scores for each pair of descriptors to supervise the learning of the detectors. It introduces the idea of backtracking to improve the localization accuracy of low-resolution feature maps. Patch2Pix~\cite{Patch2Pix} is a weakly supervised learning method that regresses pixel-level matches by predicting patch-level matches and uses the confidence scores to optimize the matching results. Although L2 normalization can bring the distribution of descriptors closer, it reduces their discriminative power. CNDesc~\cite{CNDesc} proposes to use a learnable cross-normalization technique instead of L2 normalization and designs a backbone network that effectively improves the reusability of the descriptors as well as the corresponding IDC loss to further improve the local descriptor performance.

\cite{CoAM} propose the Common Attention Module (CoAM), a spatial attention mechanism for processing two images simultaneously to determine the correspondence between them. CoAM can be embedded in other network architectures for training and applies to image datasets with significant differences. SuperGlue~\cite{SuperGlue} introduces a graph neural network and uses an attention-based context aggregation mechanism that enables it to jointly reason about matching 3D scenes. A simple but effective improvement on SuperGlue, LightGlue~\cite{LightGlueLF} reduces training difficulty and improves adaptability to difficult tasks. LoFTR~\cite{LoFTR} employs a coarse-to-fine strategy for image matching, utilizing self-attention and cross-attention layers to extract feature descriptors, and can produce dense matching pairs in low-texture regions of the image.

Multi-modal remote sensing image matching has been a research difficulty and hotspot in the field of remote sensing image matching. Firstly, there is a serious nonlinear grey scale mapping problem in multi-modal images, and secondly, the features of different modal images are independent of each other. Therefore, extracting and unifying the features of different modal images has become a key issue in matching heterogeneous remote sensing images.
There are two mainstream research directions, one for continuing to use the same steps as homologous image matching. \cite{Hughes} propose a pseudo-siamese convolution neural network architecture that puts optical and SAR images into two parallel network streams at the same time and fuses the features using a fully connected layer. \cite{FDNet} propose FDNet, which is a composite matching method consisting of a residual denoising network (RDNet) and a pseudo-siamese fully convolution network (PSFCN) with respect to optical and SAR images. CMM-Net~\cite{chaozhen} performs L2 normalization on the heterogeneous image feature descriptors to bring the heterogeneous features as close together as possible. The other converts the heterogeneous images into homogeneous images by introducing network models such as GAN before performing image matching. \cite{Zhang} and \cite{Zeng} apply image style migration to the preprocessing of image matching and eliminate radiometric and geometric differences by fusing multi-modal remote sensing images. \cite{Ma} propose a new regularised conditional generative adversarial network (GAN) for image matching preprocessing to eliminate grey scale, texture, and style differences between multi-spectral images and then use the classical local feature method to complete the heterogeneous matching task.

\begin{figure*}[t]
	\centering
	\includegraphics[width=1\linewidth]{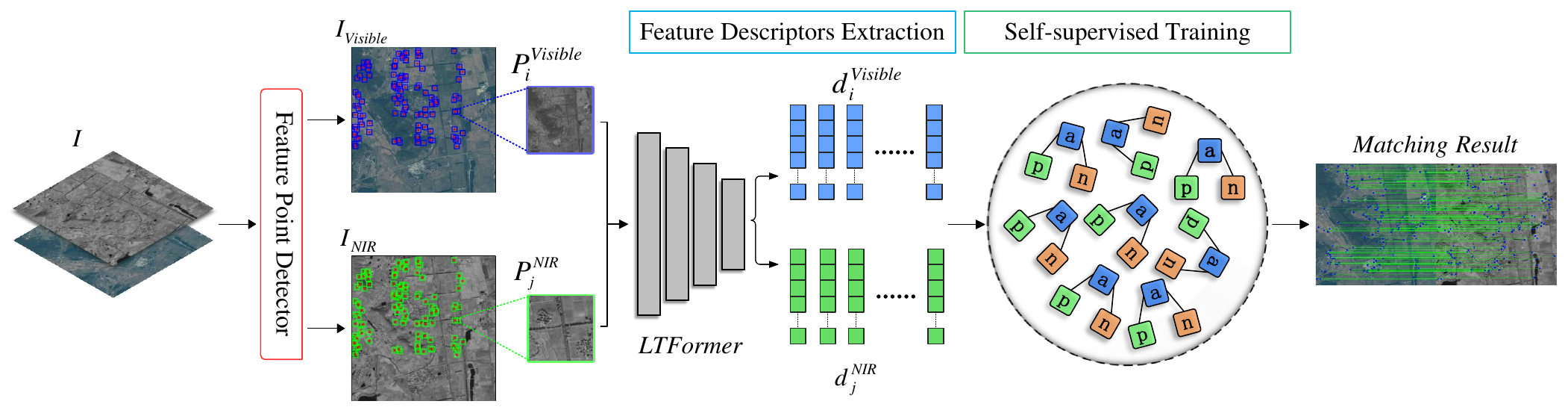}
	\caption{Overview of our LTFormer framework. The feature point detector utilizes the default SIFT algorithm, while our model is employed to generate feature descriptors. Initiating self-supervised training begins with forming a triplet of these descriptors to obtain correspondences. In conclusion, this framework facilitates the extraction of robust deep feature descriptions to match keypoints between visible and near-infrared images.}
	\label{fig:fig_framework}
\end{figure*}
\begin{figure*}[t]
	\centering
	\includegraphics[width=0.85\linewidth]{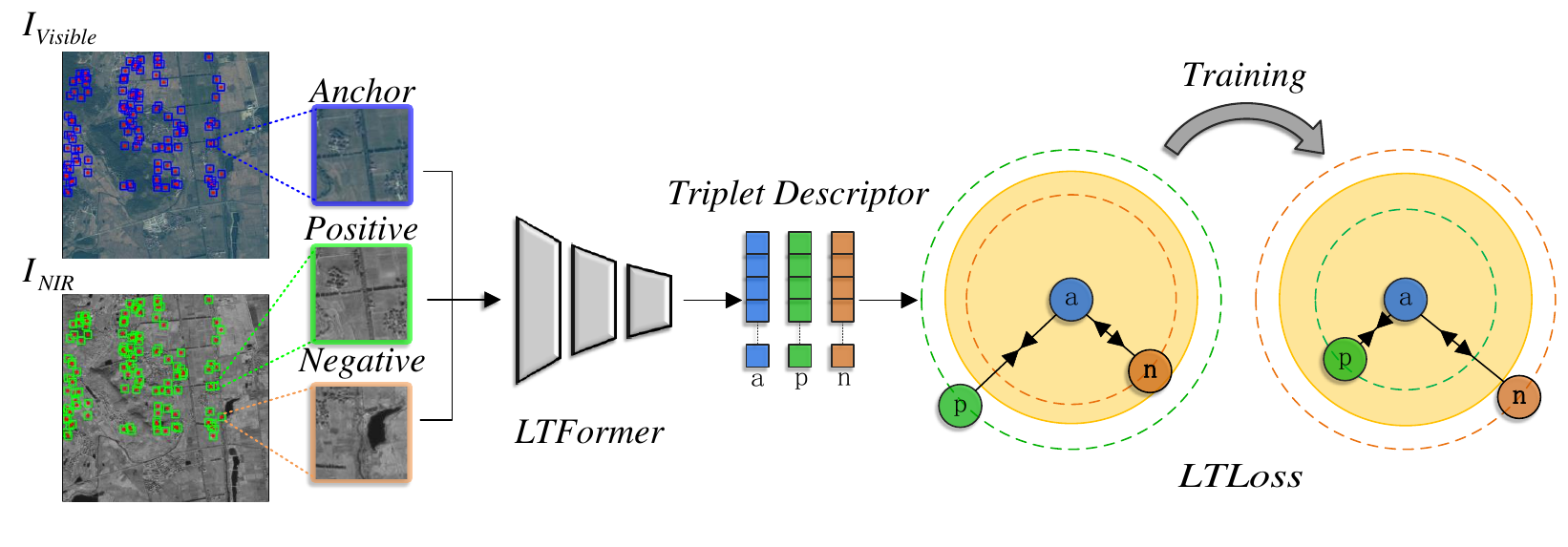}
	\caption{Overview of the self-supervised training process. By using triplet descriptors, LT Loss is utilised in the feature space to bring anchor patches as close as possible to positive patches while moving away from negative patches.}
	\label{fig:fig_self-supervised}
\end{figure*}

Because of the scarcity of labeled data in remote sensing datasets and the considerable human resources required to annotate remote sensing images with feature points, self-supervised learning eliminates the need to annotate the datasets. It saves a great deal of engineering work. FM-CycleGAN~\cite{FM-CycleGAN} introduces a feature matching loss to the unsupervised image synthesis method, CycleGAN, which aligns the features of the heterogeneous source images.

\section{Proposed Method}\label{sec:proposed}
In this section, we introduce our proposed approach for matching heterogeneous remote sensing images. First, in Section \ref{sec:overall}, we provide an overview of the general principles of the matching method used to establish correspondences among feature points in visible and near-infrared images. Next, in Section \ref{sec:self-supervised}, we describe the self-supervised training approach and the data construction strategy adapted to it. Last, we will describe in detail the various important components of the proposed approach in Section \ref{sec:model}.

\subsection{Overall Framework}\label{sec:overall}
To achieve accurate feature matching between heterogeneous remote sensing images, the key is to mitigate the impact of spectral and geometric differences and find invariant feature representation methods. As illustrated in Figure~\ref{fig:fig_framework}, here is the description of our proposed framework, which consists of four main parts: feature point detection, image patch extraction, deep feature descriptor generation, and feature matching.

Firstly, feature point detector is performed on the images to generate the center points of image patches. Assuming $I_{Visible}$ and $I_{NIR}$ are the registered visible images and near-infrared images, respectively. After applying a feature point detector, we obtain $N_{v}$ and $N_{n}$ key points on $I_{Visible}$ and $I_{NIR}$, respectively.During the image patch extraction process, we extract the surrounding regions based on the detected feature points in the visible images and near-infrared images. These regions are then reconstructed into patches of size 128$\times$128 pixels. As a result, we obtained two sets of patches: $\{P_i^{Visible}\}_{i=1:N_{v}}$ and $\{P_j^{NIR}\}_{j=1:N_{n}}$, corresponding to the visible images and near-infrared images, respectively.

To establish the correspondence between these two sets of patches, we introduce the ViT model to generate deep feature descriptors. By utilizing the PVT model, each patch is transformed into a more discriminative representation space, which improves feature utilization and reduces the number of parameters, denoted as:
\begin{equation}
	d_i^{v}=F(P_i^{Visible}), d_j^{n}=F(P_j^{NIR}),
\end{equation}
where $F(\cdot)$ denotes the ViT transformation for a given patch, $d_i^{v}$ and $d_j^{n}$ denote the feature descriptors.
Since there are significant differences between the features of different modalities, we apply L2 normalization to the descriptors of the heterogeneous image features, aiming to unify them as much as possible, denoted as:
\begin{equation}
	\hat{d_i^{v}}=d_i^r/\Vert d_i^v \Vert_2, \hat{d_j^n}=d_j^n/\Vert d_j^n \Vert_2,
\end{equation}

The normalized descriptors are then combined into a triplet form and used for self-supervised learning. By minimising the Euclidean distance between $\hat{d_i^v}$ and $\hat{d_i^n}$ in the embedding space, we establish the correspondence between $I_{Visible}$ and $I_{NIR}$, and ultimately achieve the accurate matching of feature points between visible and near-infrared images.

\subsection{Self-supervised Matching Network} \label{sec:self-supervised}
Designing a self-supervised learning paradigm is crucial for our task of generating highly discriminative models from pre-existing unannotated data. In single-modal learning, the definition of self-supervised learning relies on the training objective and whether it uses human annotations for supervision. However, in multi-modal learning, it is often possible to use one modality to provide a supervision signal for another modality. Therefore, we have adopted the concept of triplet loss and designed a self-supervised training method based on a triplet loss architecture, as shown in Figure~\ref{fig:fig_self-supervised}.

By extracting patches $\{P_i^{Visible}\}_{i=1:N_v}$ and $\{P_j^{NIR}\}_{j=1:N_n}$ from visible and near-infrared images, we reconstruct triplet image patches suitable for self-supervised training as inputs, denoted as $(P_i,P_i^+,P_i^-)$. $P_i$ represents the anchor patch, $P_i^+$ represents the positive patch, and $P_i^-$ represents the negative patch. We can take a given visible image patch as an anchor patch Pi and apply a homography transformation $T(\cdot)$ to the corresponding near-infrared image patch to generate a positive patch $P_i^+$, and a negative patch $P_i^-$ can be any other patch from the same set of near-infrared image patches. The homography transformation $T(\cdot)$ generates copies of images with different appearances, which helps the model learn feature representations with invariance and improves the model's generalization ability. By choosing appropriate positive and negative patches, we can form them into a triad for input into the network and obtain the corresponding deep feature descriptors. The goal of training is to minimize the distance between the anchor patch $P_i$ and the positive patch $P_i^+$ while maximizing the distance between the anchor patch $P_i$ and the negative patch $P_i^-$.

\begin{table}[t]
	\renewcommand\arraystretch{1.2}
	\centering
	\caption{Description of the proposed LTFormer architecture.}
	\label{tab:modelsetting}
	\tiny
	\setlength{\tabcolsep}{3.5pt}
	\resizebox{\columnwidth}{!}{
		\begin{tabular}{c|c|c|c}
			\Xhline{1.2pt}
			&\textbf{Output Size}&\textbf{Layer Name} &\textbf{\makecell[c]{Light-weight \\ Transformer}} \\
			\Xhline{1.2pt}
			\multirow{6}{*}{\textbf{Stage1}}&\multirow{6}{*}{{$\frac{H}{4} \times \frac{W}{4} $}}&\multirow{2}{*}{\makecell[c]{Overlapping Patch Embedding}} &$S_1=4$ \\
			&&&$C_1 = 16$ \\
			\cline{3-4}
			&&\multirow{4}{*}{Transformer Encoder}&$R_1=8$ \\
			&& &$N_1=1$  \\
			&& &$E_1=8$  \\
			&& &$L_1=2$  \\
			\hline
			
			\multirow{6}{*}{\textbf{Stage2}}&\multirow{6}{*}{{$\frac{H}{8} \times \frac{W}{8} $}}&\multirow{2}{*}{Overlapping Patch Embedding} &$S_2=2$ \\
			&&&$C_2 = 32$ \\
			\cline{3-4}
			&&\multirow{4}{*}{Transformer Encoder}&$R_2=4$  \\
			&& &$N_2=2$  \\
			&& &$E_2=8$  \\
			&& &$L_2=2$  \\
			\hline
			
			\multirow{6}{*}{\textbf{Stage3}}&\multirow{6}{*}{{$\frac{H}{16} \times \frac{W}{16} $}}&\multirow{2}{*}{Overlapping Patch Embedding} &$S_3=2$\\
			&&&$C_3 = 64$ \\
			\cline{3-4}
			&&\multirow{4}{*}{Transformer Encoder}&$R_3=2$ \\
			&& &$N_3=4$  \\
			&& &$E_3=4$ \\
			&& &$L_3=2$  \\
			\hline
			
			\multirow{6}{*}{\textbf{Stage4}}&\multirow{6}{*}{{$\frac{H}{32} \times \frac{W}{32} $}}&\multirow{2}{*}{Overlapping Patch Embedding} &$S_4=2$ \\
			&&&$C_4 = 128$ \\
			\cline{3-4}
			&&\multirow{4}{*}{Transformer Encoder}&$R_4=1$ \\
			&& &$N_4=8$  \\
			&& &$E_1=8$  \\
			&& &$L_4=2$  \\
			\Xhline{1.2pt}
	\end{tabular}}
\end{table}

\subsection{Light-weight Transformer and Loss Function}\label{sec:model}
Inspired by PVT (Pyramid Vision Transformer), we further streamline the network model based on PVTv2, and finally get the version named Light-weight Transformer (LTFormer). Although PVTv2 has outperformed Swin-B and has fewer parameters and computational effort, it still requires a large amount of computational resources when dealing with high-resolution remote sensing image tasks. In order to achieve the goal of lightweight, we reduce the number of channels in the PVTv2-B0 network from the original $C_i=\{32, 64, 160, 256\}_{i=1:4}$ to  $C_i=\{16, 32, 64, 128\}_{i=1:4}$. The specific hyperparameter settings are shown in the following table.

In the context of remote sensing imagery, the interference of factors such as scale variation and viewpoint change poses challenges to feature extraction. Additionally, images from different sensors can exhibit significant differences in spectral and geometric properties, even under unchanged viewpoints. Therefore, in the task of feature extraction from remote sensing images, it becomes crucial to employ adversarial sample training and to seek a more discriminative feature space. The Triplet Loss\cite{TripletLoss}, commonly used in face recognition tasks, can provide adversarial signals and facilitate the model in learning the differences between samples without relying on label information, thereby meeting the requirements of self-supervised training.

\begin{figure*}[h]
	\centering
	\includegraphics[width=1\linewidth]{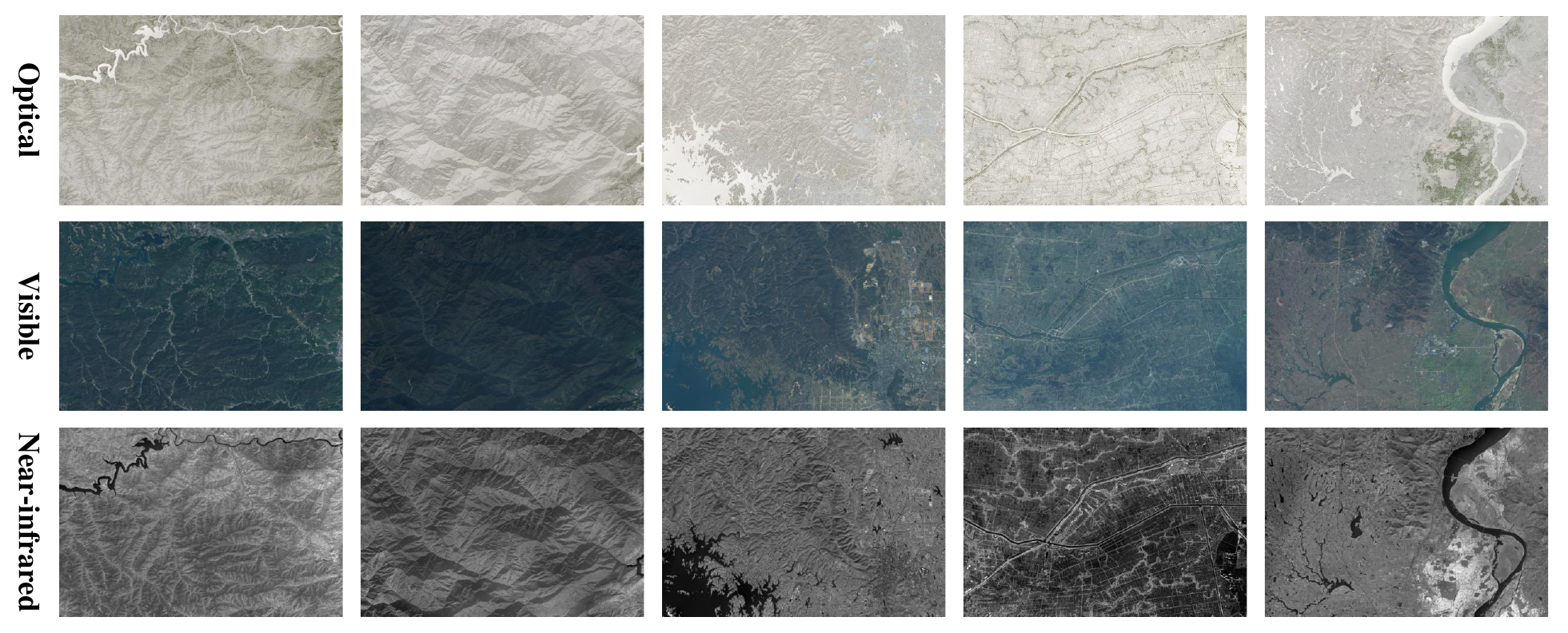}
	\caption{Samples of The WHU-OPT-SAR dataset. The optical image is obtained by merging a visible image with a near-infrared image.}
	\label{fig:fig_dataset}
\end{figure*}

In this case, our training samples consist of N sets of three 128-dimensional patch embeddings $(P_i,P_i^+,P_i^-)$, forming triplets to compute the distance differences between samples of the same class and different classes. The learning objective of the loss function is to increase the distance between different class samples while reducing the distance between samples of the same class, ensuring that the distance between samples of the same class is smaller than that between samples of different classes. The specific formulation of the loss function is as follows:
\begin{equation}
	LT Loss = \frac{1}{N_{v}}\sum_{i=0}^{N_{v}} max\{D(P_i,P_i^+)+D(P_i,P_i^-)-M , 0 \},
\end{equation}
where $D(P_i,P_i^+)$ denotes the calculation of the Euclidean distance between $P_i$ and $P_i^+$, $M$ is a boundary value that controls the difference in distance between samples of the same class and samples of different classes, and the value is determined as
\begin{equation}
	M = |D(P_i,P_i^+)+D(P_i,P_i^-)|/2.
\end{equation}
By adaptively adjusting the boundary value $m$ according to the distance difference between samples, it can help the model to better learn the differences between samples.

\section{Experiment and Analysis}\label{sec:experiment}
\subsection{Experiment Setup}
\subsubsection{Database description} 

The experiment utilizes the WHU-OPT-SAR dataset~\cite{li2022mcanet}, which consists of images with four channels: red (R), green (G), blue (B), and near-infrared (NIR). The dataset covers an area of 51,448.56 square kilometers with a resolution of 5 meters. It includes diverse remote sensing images of different terrains such as mountains, forests, hills, and plains, as well as various vegetation types, including coniferous forests, broad-leaved forests, shrubs, and aquatic vegetation, as shown in Figure \ref{fig:fig_dataset}. The WHU-OPT-SAR dataset comprises a total of 100 optical images with dimensions of 5556$\times$3704 pixels. The first 80 images are used as the training set, while the remaining 20 images serve as the validation set.

\begin{figure*}[t]
	\centering
	\includegraphics[width=1\linewidth]{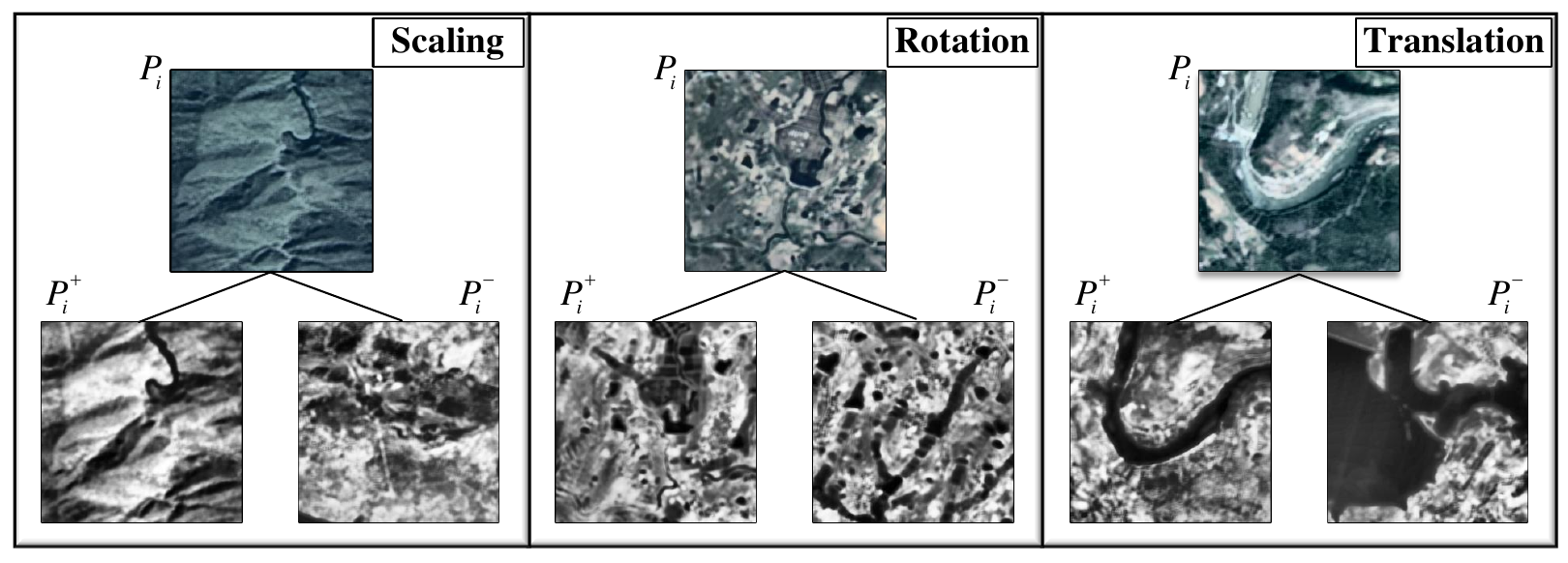}
	\caption{Sample triplex dataset showing triplex morphology after different homography transformations.}
	\label{fig:fig_triplet}
\end{figure*}

The initial step involves separating the near-infrared channel from the original images and dividing the dataset into visible and near-infrared data groups. Due to the high resolution of the original images, which hinders feature point detection, we crop the images to a size of 926$\times$926 pixels. In order to maintain the same number of channels as the NIR image, we convert the visible image to a grayscale image. Then, we apply Contrast Limited Adaptive Histogram Equalization (CLAHE) to enhance the contrast of the images. Following the guidelines in section \ref{sec:model}, we extract a large number of feature points from appropriately sized images and subject them to anchor-based cropping. To generate positive samples, we apply appropriate transformations to the near-infrared image patches. The experiment provides three transformation methods: small-scale scaling (Sf = [0.9, 0.95, 1.05, 1.1, 1.15]), small-angle rotation ($\theta$ $\in$ [5, 10, 15] degrees), and small-scale translation (8 pixels). Combining all the steps, we create a training dataset consisting of 10,000 groups of patches, as shown in Figure \ref{fig:fig_triplet}.

\subsubsection{Evaluation Metrics} 
The experiment employs two metrics to evaluate the algorithm performance: Matching Precision and Matching Score. Matching Precision is defined as the ratio of the number of correctly matched points ($N_{corr}$) to the number of key points successfully matched ($N_{suss}$). Matching Score is defined as the ratio of the number of correctly matched points to the total number of matched points. The determination of correctly matched points can be performed using the following formula:
\begin{equation}
	Corr(x):\Vert x_i - x_j\Vert \le \epsilon
\end{equation}
where $x_i$ represents the key points matched by the algorithm, and $x_j$ represents the ground truth key points. They are considered correct match points if the Euclidean distance between them is less than the given projection error $\epsilon$. The projection error is fixed at $\epsilon = 5 $ in our experiment.

$N_{suss}$ depends on the matching threshold, which measures the Euclidean distance between the depth feature descriptors the algorithm determines to be two matches. A matching threshold 0.5 is used, and any distance greater than this threshold is considered a failed match point.

The algorithm's performance on the feature point matching task can be evaluated by calculating the matching accuracy and matching score. The matching accuracy reflects the accuracy of the algorithm. In contrast, the matching score considers both the accuracy and recall of the algorithm, providing a more comprehensive assessment of the algorithm's performance.

\subsubsection{Implementation Details} 
The experiment is conducted without pre-trained weights and based on self-supervised learning. It utilizes the Stochastic Gradient Descent (SGD) optimizer for training, with a batch size 256. The initial learning rate is set to 0.001, and a momentum of 0.9 is applied. The training is performed on an NVIDIA A100 GPU with 40GB of memory. It consists of 50 epochs and takes 2 hours to complete. The dimension of the deep feature descriptor is set to 128 by default, the input image size of the model is set to 128$\times$128 pixels, the feature extractor is set to SIFT by default, and the patch size is set to 64. The matching time for a set of image pairs with dimensions of 463$\times$463 pixels is 1 second.

\subsection{Comparison to state-of-the-art local feature descriptor methods}
Due to the confidentiality and limited accessibility of remote sensing images, there is currently no publicly available benchmark dataset for evaluating the performance of remote sensing image matching. Therefore, we need to create an annotated validation dataset to perform the evaluation.
We are currently selecting the last 20 images from the WHU-OPT-SAR dataset and dividing them into visible light and NIR groups. Each image is cropped to a size of 463$\times$463 pixels, and the coordinates of feature points are recorded using a feature extractor. Since the NIR images are extracted from the original dataset, each visible-NIR image pair is aligned, and the feature points correspond one-to-one. This process results in a validation dataset consisting of 1940 image pairs.

\subsubsection{Comparison to state-of-the-art Handcrafted descriptors} 
\begin{figure*}[t]
	\centering
	\includegraphics[width=0.9\linewidth]{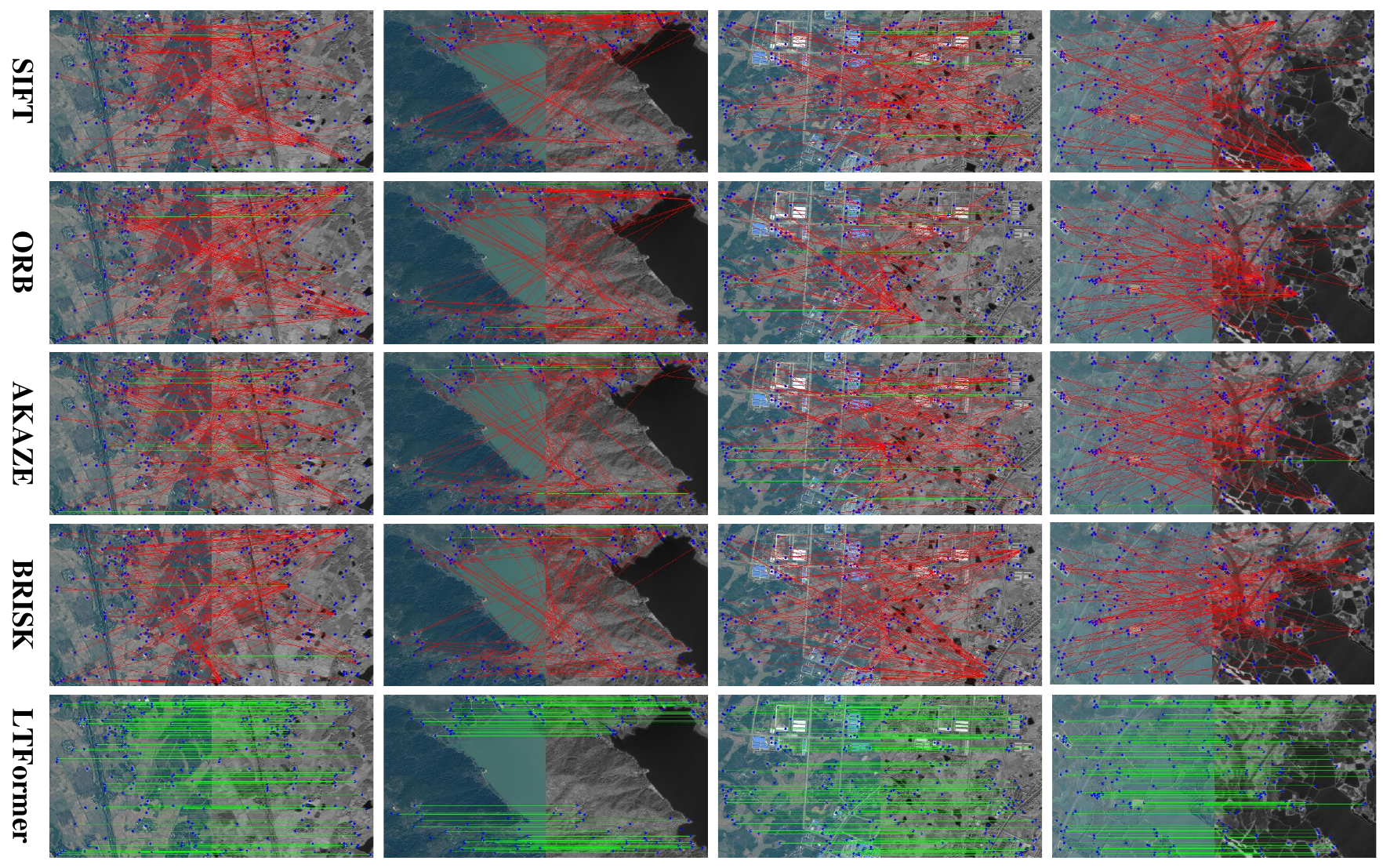}
	\caption{Qualitative comparison of LTFormer with traditional methods such as SIFT, ORB, AKAZE and BRISK.}
	\label{fig:fig_compare}
\end{figure*}

To fairly compare the performance of different descriptors, we compare traditional methods (including SIFT, ORB, AKAZE, and BRISK) and our proposed LTFormer descriptor on the same set of keypoints. Figure~\ref{fig:fig_compare} displays the matching results between visible and near-infrared images. Green lines represent correct matches, while incorrect matches are shown in red. It is evident from the results that the LTFormer descriptor achieves a significantly higher number of correct matches compared to the traditional descriptors. This indicates a clear advantage of LTFormer in matching heterogeneous images.

\begin{table}[t]
	\renewcommand\arraystretch{1}
	\centering
	\caption{Quantitative comparison of LTForme with deep learning-based feature descriptors (e.g. SOSNet, SuperGlue, HyNet, CNDesc, and HardNet) at different patch size scales.}
	\label{tab:comperelearning}
	\setlength{\tabcolsep}{3.5pt}
	\resizebox{0.95\columnwidth}{!}{
		\begin{tabular}{clcc}
			\Xhline{1.2pt}
			\textbf{Patch Size} &\textbf{Method} &\textbf{Precision} &\textbf{Matching Score} \\
			\Xhline{1.2pt}
			\multirow{6}{*}{64}  &SoSNet~\cite{SosNet} &0.8459 &0.7389 \\
			&SuperGlue~\cite{SuperGlue} &0.8381 &0.7511 \\
			&HyNet~\cite{HyNet} &0.8446 &0.7462 \\
			&CNDesc~\cite{CNDesc} &0.8302 &0.7284 \\
			&HardNet~\cite{HardNet} &0.8453 &0.7485\\
			&Proposed &\textbf{1.0000} &\textbf{1.0000}\\
			\hline
			\multirow{6}{*}{128}  &SoSNet~\cite{SosNet} &0.9873 &0.9363 \\
			&SuperGlue~\cite{SuperGlue} &0.9807 &0.9484 \\
			&HyNet~\cite{HyNet} &0.9886 &0.9232 \\
			&CNDesc~\cite{CNDesc} &0.9727 &0.9254 \\
			&HardNet~\cite{HardNet} &0.9879 &0.9459\\
			&Proposed  &\textbf{1.0000} &\textbf{1.0000}\\
			\hline
			\multirow{6}{*}{192}  &SoSNet~\cite{SosNet} &0.9881 &0.9592 \\
			&SuperGlue~\cite{SuperGlue} &0.9815 &0.9702 \\
			&HyNet~\cite{HyNet} &0.9894 &0.9452 \\
			&CNDesc~\cite{CNDesc} &0.9737 &0.9481 \\
			&HardNet~\cite{HardNet} &0.9887 &0.9682\\
			&Proposed  &\textbf{1.0000} &\textbf{1.0000}\\
			\Xhline{1.2pt}
	\end{tabular}}
\end{table}

\subsubsection{Comparison to state-of-the-art learning feature descriptors}
Although there is no benchmark test specifically for keypoint matching between visible and near-infrared images, we conduct our own experiments and compare our results with the latest deep learning-based keypoint matching methods. We construct a dataset by ourselves and divide it into a training set and a validation set with a 4:1 ratio. We perform benchmark tests on supervised learning methods such as SosNet, SuperGlue, HyNet, and CNDesc, as well as self-supervised learning methods like HardNet~\cite{HardNet}. In the experiments, we use the SIFT method as the feature point extractor to ensure that all methods were trained based on the same set of keypoints. Table \ref{tab:comperelearning} summarizes the matching accuracy and matching score results obtained by the six methods at different scales. It can be observed that our method consistently achieved the best results in terms of matching accuracy and matching scores across different image scales.

\subsection{Ablation Studies and Analysis}

\subsubsection{Comparison of Feature Point Detector}
\begin{figure*}[t]
	\centering
	\includegraphics[width=0.75\linewidth]{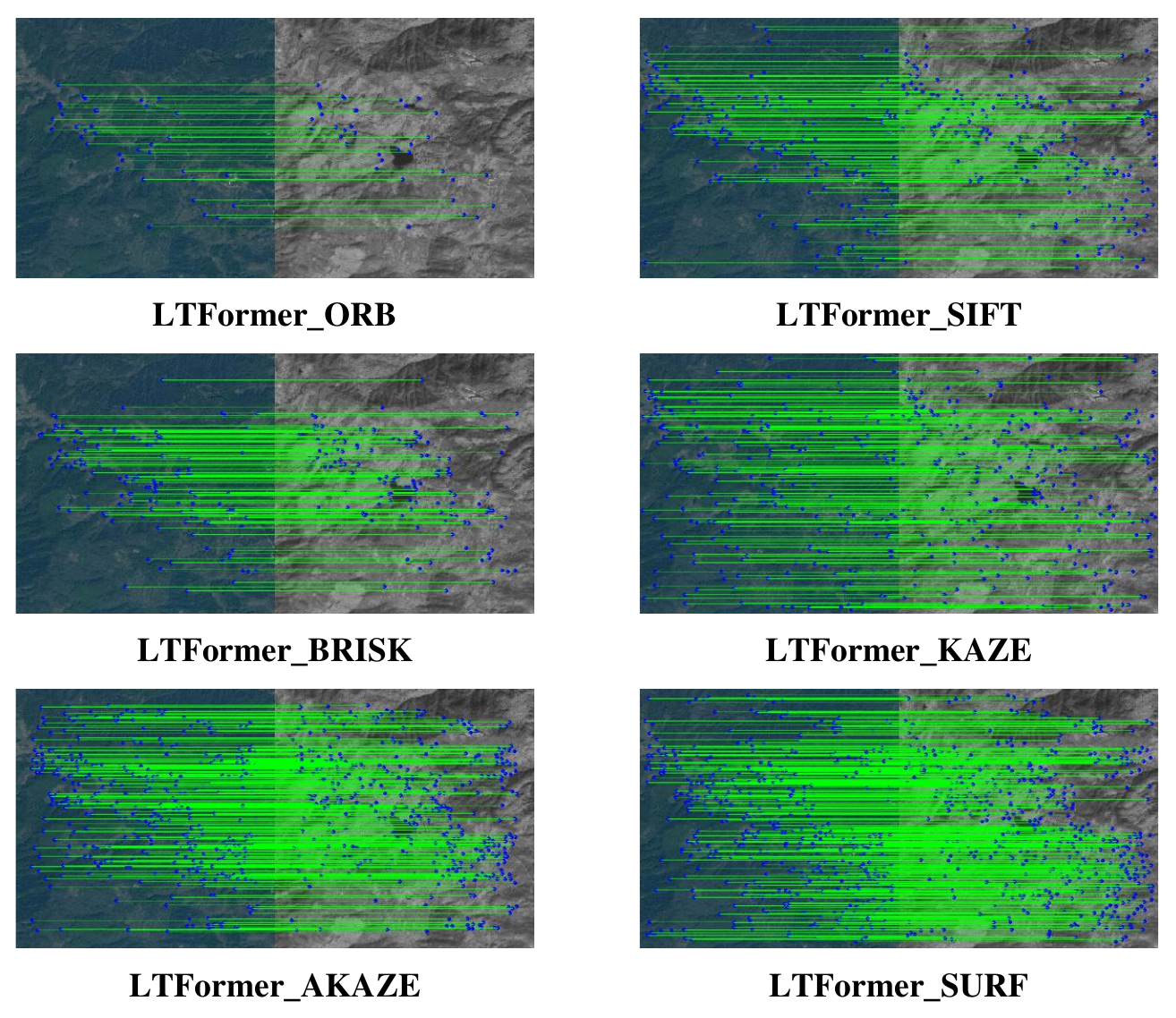}
	\caption{Matching results of LTFormer descriptors on SIFT, SURF, AKAZE, KAZE, ORB and BRISK feature point Detectors.}
	\label{fig:fig_feature}
\end{figure*}

We conduct LTFormer descriptor testing on six feature point extractors: SIFT, SURF, AKAZE, KAZE, ORB, and BRISK. We successfully match all feature points using these extractors, as shown in Figure~\ref{fig:fig_feature}. This indicates that our method achieves favorable results on traditional feature point extractors like SIFT and SURF and updated ones such as AKAZE, KAZE, ORB, and BRISK. The results demonstrate that the LTFormer descriptor exhibits broad applicability and robustness, enabling accurate feature point matching across various feature point extractors. This powerful tool can be utilized for tasks such as image registration, object detection, and image retrieval.

\subsubsection{Comparison of triplet loss variants}
We compare different triplet loss functions, including HardNet Loss \cite{HardNetLoss}, standard Triplet Loss with a fixed margin \cite{TripletLoss}, adaptive margin triplet loss \cite{AdaptativeLoss}, and our proposed LT Loss. Based on the comparison results (see Table \ref{tab:loss}), we draw the following conclusions:

From the metrics, HardNet Loss performs poorly on heterogeneous image datasets with large inter-class differences and intra-class similarities. The margin selection schemes of standard Triplet Loss with a fixed margin and adaptive margin triplet loss also demonstrate their limitations. This suggests that these triplet loss functions may not adequately meet the distribution characteristics of the samples, thereby affecting the performance of feature learning. In contrast, our designed LT Loss exhibits excellent performance. This indicates that our loss function possesses higher discriminability, robustness, and adaptability.

\begin{table}[t]
	\renewcommand\arraystretch{1.2}
	\centering
	\caption{Comparison of four different loss functions: Adaptive margin triple Loss, HardNet Loss, Triple Loss and LT Loss in terms of accuracy and score matching. LT Loss gives better performance.}
	\label{tab:loss}
	\setlength{\tabcolsep}{3.5pt}
	\resizebox{0.85\columnwidth}{!}{
		\begin{tabular}{lcc}
			\Xhline{1.2pt}
			\textbf{Loss Function}&\textbf{Precision} &\textbf{Matching Score} \\
			\hline
			Adaptative Loss\cite{AdaptativeLoss} &0.9883 &0.9263   \\
			HardNet Loss\cite{HardNetLoss}  &0.9886 &0.9507 \\
			Triplet Loss\cite{TripletLoss} &0.9915 &0.9915 \\
			LT Loss &\textbf{1.0000} &\textbf{1.0000} \\
			\Xhline{1.2pt}
	\end{tabular}}
\end{table}

\begin{table}[!tbp]
	\renewcommand\arraystretch{1.2}
	\centering
	\caption{Performance of LTFormer on different feature dimensions.}
	\label{tab:Dim}
	\tiny
	\setlength{\tabcolsep}{3.5pt}
	\resizebox{0.85\columnwidth}{!}{
		\begin{tabular}{lcc}
			\Xhline{0.85pt}
			\textbf{Descriptor Dimension}&\textbf{Precision} &\textbf{Matching Score} \\
			\hline
			64 &0.9998&0.9998 \\
			128 &\textbf{1.0000}&\textbf{1.0000} \\
			256 &\textbf{1.0000}&\textbf{1.0000} \\
			\Xhline{0.8pt}
	\end{tabular}}
\end{table}

\subsubsection{Comparison of feature descriptor dimension settings}
The dimension of a feature descriptor significantly impacts its descriptive capability, particularly in the context of deep learning-based descriptors. Higher-dimension descriptors can provide richer feature representations by capturing more intricate details and variations in the data. However, they often come with increased computational costs. On the other hand, reducing the dimension of a feature descriptor can lower the computational complexity. Still, it may also result in a less discriminative representation with a reduced ability to capture fine-grained details. It's essential to balance the dimension and the desired level of descriptive power, considering the specific application requirements and available computational resources. For example, in our experiments with LTFormer descriptors, we test different dimensions, as shown in Table \ref{tab:Dim}. Even at the lowest dimension of 64, the LTFormer descriptors maintain excellent performance comparable to descriptors with higher dimensions, such as 128 or 256. This indicates that LTFormer can effectively capture feature information with fewer parameters and computational requirements, making it well-suited for practical applications in remote sensing image matching.

\section{Conclusion} \label{sec:conclusions}
This paper proposes a self-supervised matching network termed as LTFormer, based on the light-weight transformer architecture, to address the challenge of matching heterogeneous remote sensing images. Leveraging a triplet input and a well-defined data construction strategy, visible and near-infrared images are input into the matching network to derive LTFormer descriptors. Additionally, the LT Loss function is incorporated to enhance correspondence search within a more discriminative feature space. Notably, our self-supervised approach necessitates no annotated data, relying solely on the original heterogeneous remote sensing images for training. Comparisons with traditional handcrafted feature descriptors and recent supervised deep learning methods demonstrate that our approach outperforms other methods significantly in feature matching. Future research directions include exploring more optimal network structures and loss functions to further improve the performance of matching heterogeneous remote sensing images.

\bibliographystyle{IEEEtran}
\bibliography{references}

\end{document}